\documentclass[11pt]{article}

\usepackage[final]{acl}

\usepackage{times}
\usepackage{latexsym}

\usepackage[T1]{fontenc}

\usepackage[utf8]{inputenc}

\usepackage{microtype}

\usepackage{inconsolata}

\usepackage{graphicx}

\graphicspath{{figures/}}

\title{Hospitality-VQA: Decision-Oriented Informativeness Evaluation for Vision–Language Models}

\author{
  \textbf{Jeongwoo Lee\textsuperscript{1,$\ast$}},
  \textbf{Duhyeong Baek\textsuperscript{1}},
  \textbf{Eungyeol Han\textsuperscript{1}},
  \textbf{Soyeon Shin\textsuperscript{1}},
\\
  \textbf{Gukin Han\textsuperscript{2}},
  \textbf{Seungduk Kim\textsuperscript{2}},
  \textbf{Jaehyun Jeon\textsuperscript{1,$\dagger$}},
  \textbf{Taewoo Jeong\textsuperscript{2,$\dagger$}}
\\
  \textsuperscript{1}\texttt{\{leejeongwoo9941,glzeng99,condense,shin020810,jaehyun.jeon\}@yonsei.ac.kr}
\\
  \textsuperscript{2}\texttt{\{bryan.han,seungduk.kim,taewoo.jeong\}@yanolja.com}
\\
\\
  \textsuperscript{1}Yonsei University,
  \textsuperscript{2}Yanolja NEXT
\\
}

\begin{document}
\maketitle

\begingroup
\renewcommand\thefootnote{}
\footnotetext{$\ast$\ \ Main\ contributor.}
\footnotetext{$\dagger$\ \ Corresponding\ author.}
\endgroup

\begin{abstract}
Recent advances in Vision–Language Models (VLMs) have demonstrated impressive multimodal understanding in general domains. 
However, their applicability to decision-oriented domains such as hospitality remains largely unexplored. 
In this work, we investigate how well VLMs can perform visual question answering (VQA) about hotel and facility images that are central to consumer decision-making. While many existing VQA benchmarks focus on factual correctness, they rarely capture what information users actually find useful. To address this, we first introduce \textit{Informativeness} as a formal framework to quantify how much hospitality-relevant information an image–question pair provides.
Guided by this framework, we construct a new hospitality-specific VQA dataset that covers various facility types, where questions are specifically designed to reflect key user information needs. Using this benchmark, we conduct experiments with several state-of-the-art VLMs, revealing that VLMs are not intrinsically decision-aware—key visual signals remain underutilized, and reliable informativeness reasoning emerges only after modest domain-specific finetuning.
\end{abstract}

\section{Introduction}

\begin{figure}[t]
    \centering
    \includegraphics[width=\columnwidth]{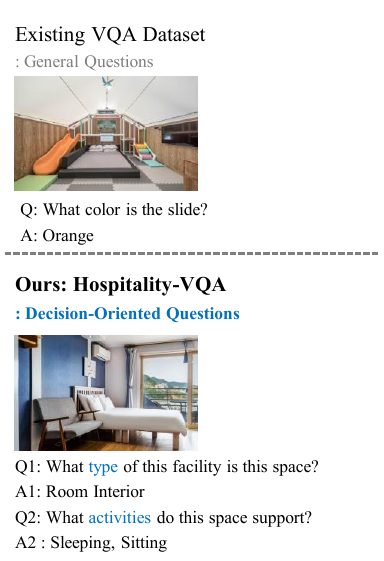}
    \caption{
    Comparison between general VQA (top) and decision-oriented Hospitality-VQA (bottom).
    }
    \label{fig:fig1}
\end{figure}

Images play a central role in the hospitality industry, serving as the primary medium through which guests evaluate and compare accommodations~\citep{zhang2022makes}.
When consumers choose where to stay, they often rely more on visual impressions—such as room layout, view, lighting, and cleanliness—than on textual descriptions. 
These images convey both factual and atmospheric cues that shape user decisions, making visual understanding a crucial aspect of hospitality intelligence~\citep{CuestaValino2023Aesthetics}.

Previous studies in the hospitality domain have predominantly relied on text-based analytics of online reviews to model customer satisfaction, preferences, and demand patterns~\citep{li2013determinants, xiang2015can}. In parallel, a growing body of work has examined visual information in accommodation images by extracting predefined or low-level features—such as aesthetics, composition, or object categories—and relating them to outcomes such as booking decisions, user intentions, or perceived accommodation quality~\citep{ren2021large, he2023image}. More recently, while some studies have leveraged Large Language Models (LLMs) for hospitality analysis, they remain primarily focused on textual inputs such as reviews or descriptions~\citep{guidotti2025discovering}. Despite these advancements, existing methods—whether text-centric or feature-based—remain limited in their ability to perform integrated multimodal reasoning. Specifically, they often fail to capture the interplay between higher-level spatial organization and functional semantics in images, factors that are central to how humans evaluate hospitality environments.

Meanwhile, recent advances in Vision-Language Models (VLMs) have significantly improved multimodal reasoning across general domains. Modern models~\citep{comanici2025gemini,hurst2024gpt4o,bai2025qwen25vltechnicalreport} can generate contextualized image descriptions and answer open-ended questions that go beyond traditional visual recognition, suggesting strong potential for domain-specific applications. While these models have demonstrated promising results in specialized fields such as e-commerce~\citep{trabelsi2025product} and medical imaging~\citep{tu2024towards}, their use in the hospitality domain has been relatively limited with respect to decision-oriented evaluation settings.

When examining these domain-specific applications, one important insight emerges: the performance of VLMs often depends on \textit{how information needs are framed}. Generic prompts (e.g., "What is in this image?") yield vague descriptions that are insufficient for hospitality purposes.
As illustrated in Figure~\ref{fig:fig1}, appearance-level questions alone provide limited insight into whether a space meaningfully supports guest activities or experiences.
Meaningful evaluation requires domain-specific questions that elicit decision-relevant insights---\textit{not just whether a room contains furniture, but how its layout supports guest activities; not merely whether a window exists, but what type of view it provides}.
This raises a key design challenge: how to formalize the kinds of visual evidence that actually support user decisions.

To address this challenge, we introduce \textbf{Hospitality Informativeness}, a domain-grounded framework that quantifies how much decision-relevant information a hospitality image–question pair provides.
Because user information needs vary across facility types—such as spatial clarity in rooms, amenity completeness in bathrooms, or functional elements in shared facilities~\citep{wakefield1996servicescape}—we first identify the facility type and design domain-specific questions accordingly.
Although these needs appear diverse, we observe that the visual cues influencing booking decisions consistently fall into a small set of structural, functional, and view-related dimensions.
Building on this observation, we define four fundamental visual axes (spatial legibility, activity affordance, contextual openness, and geometric completeness).
Together, these axes capture the dominant cues that shape user perception and decision-making in hospitality imagery, providing a principled basis for evaluating VLM responses~\citep{Greene2016Function}. 
We use these axes to construct \textbf{Hospitality-VQA}, a new VQA benchmark aligned with decision-centric evaluation rather than generic scene description.

Our main contributions are:
\begin{itemize}
    \item We formalize \emph{Informativeness} in the hospitality domain as a set of four interpretable axes that capture decision-relevant visual cues in hotel and facility imagery.
    \item We build \textbf{Hospitality-VQA}, a VQA dataset whose questions and labels are derived from these axes and tailored to diverse facility types.
    \item We benchmark eight general-purpose VLMs and show that they struggle with fine-grained hospitality informativeness. Our dataset enables measurable performance gains through lightweight domain adaptation, highlighting its value as a foundation for future model development on hospitality domain.
\end{itemize}

\begin{figure*}[!htbp]
    \centering
    \includegraphics[width=\textwidth]{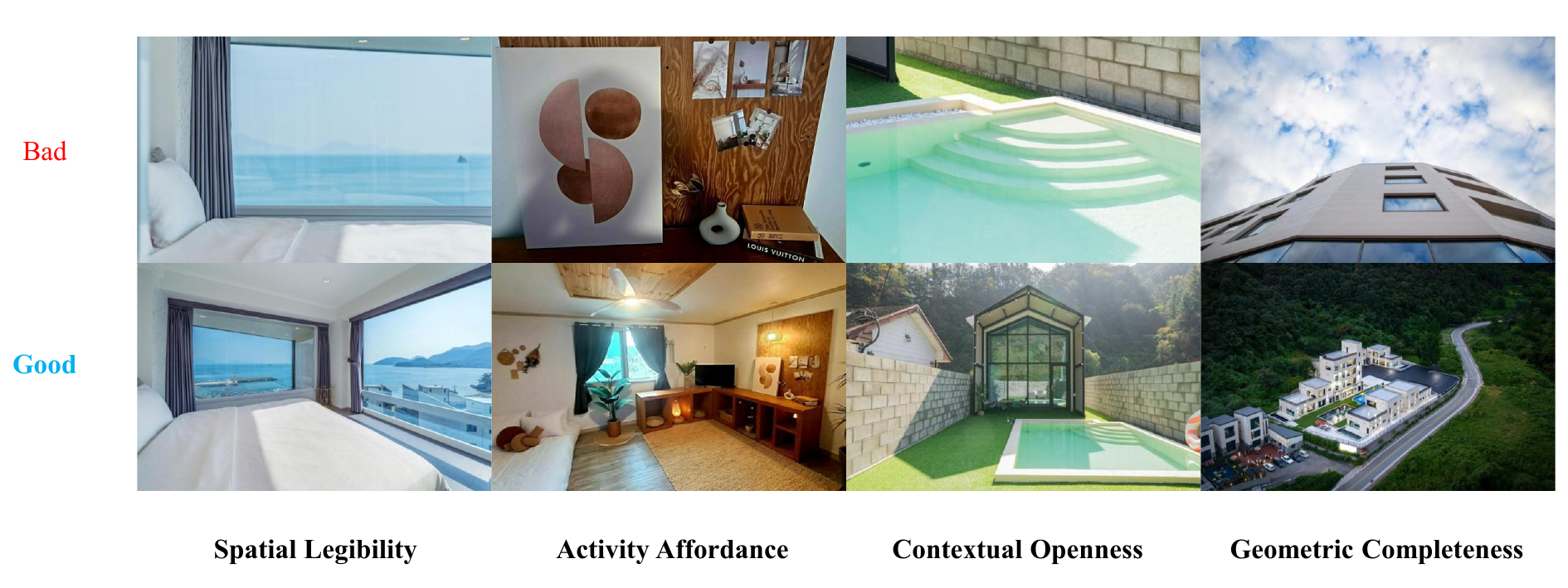}
    \caption{Bad vs. Good examples for each informativeness dimension. Bad images lack decision-relevant visual cues—resulting in low  spatial legibility, weak activity affordance, obstructed or unbalanced contextual openness, or incomplete geometric completeness. Good images exhibit high spatial legibility, clear activity affordances, well-balanced contextual openness, and strong geometric completeness, enabling more reliable assessment of hospitality informativeness.}
    \label{fig:informativeness_axis}
\end{figure*}

\section{Related Works}

\subsection{Visual Analysis in Hospitality}


Research in hospitality AI has largely focused on structured prediction tasks such as room-type classification and price estimation using CNN-based frameworks, treating images as static inputs and overlooking richer semantic cues relevant to user assessment. 
In parallel, a growing line of work extracts computable visual descriptors—ranging from low-level color statistics to mid-level attributes such as aesthetics and composition—and relates them to outcomes like booking intentions or demand~\citep{zhang2022makes, he2023image, CuestaValino2023Aesthetics}. 
However, these approaches are not designed to evaluate whether models can answer \emph{decision-relevant questions} about an image. 
Recent work has also explored multimodal hotel retrieval and preference matching~\citep{askari2025hotelmatch}, but focuses on similarity or relevance rather than explicit question answering and decision-oriented evaluation.

Existing approaches rarely model how guests simulate a potential stay experience from visual evidence. 
Although the presentation of accommodation photographs can sway selection behavior~\citep{SanchezTorres2024Visual}, the notion of visual utility—how visual elements convey functional and spatial affordances—remains under-specified. 
Consequently, evaluation typically centers on prediction accuracy or correlational signals rather than decision-oriented reasoning. 
Our work addresses this gap by shifting the focus to the systematic evaluation of decision-relevant information through a VQA benchmark grounded in \emph{Hospitality Informativeness}.

\subsection{Vision–Language Models and Domain Adaptation}

Recent general-purpose VLMs, including GPT-4o~\citep{hurst2024gpt4o} and Gemini 2.5 Pro~\citep{comanici2025gemini}, demonstrate impressive capabilities in image captioning and open-ended QA. However, these models are trained primarily on web-scale, caption-style data to describe “what exists,” often lacking the specialized reasoning required to evaluate “how useful it is” in a vertical domain. In hospitality, visual understanding goes beyond object recognition; it requires inferring spatial habitability and functional affordance. Since standard VLMs are not inherently optimized for such evaluative reasoning, it remains unclear to what extent they can interpret the nuanced visual evidence essential for consumers—motivating the need for a domain-grounded benchmark.

\subsection{From Factuality to Decision-Centric VQA}

Standard VQA benchmarks (e.g., VQA v2~\citep{goyal2017making}, GQA~\citep{hudson2019gqa}) have driven progress in multimodal reasoning but primarily evaluate factual correctness or commonsense knowledge. While recent goal-oriented VQA tasks explore navigation or physical manipulation~\citep{das2018embodied,gurari2018vizwiz}, they rarely address consumer-facing decisions in which the goal is to assess the suitability of a space or service. Existing benchmarks are not designed to measure whether an image provides the type of evidence needed to support informed accommodation choices~\citep{CuestaValino2023Aesthetics}. Addressing this limitation, we introduce \textit{Informativeness} as a metric to quantify the specific visual signals—such as layout clarity and functional completeness—that facilitate reliable accommodation assessment.

\section{Quantifying Informativeness in Hospitality}
\label{sec:info}

We argue that true understanding in the hospitality domain requires quantifying the visual evidence that supports user decision-making. While general VQA benchmarks focus on factual correctness (e.g., “is there a window?”), hospitality users rely on images to envision their stay—judging layout, affordance, and atmosphere. Because these subjective assessments directly drive booking decisions, mere descriptions are insufficient~\citep{CuestaValino2023Aesthetics}. To address this, we formalize \textit{Informativeness} as a measurable metric. We propose that the vague notion of “a useful hotel image” can be decomposed into specific, quantifiable axes that act as proxies for the user’s envisioned stay experience~\citep{Greene2016Function}.

\subsection{Facility Taxonomy and Informativeness Dimensions}

Hospitality imagery encompasses diverse scenes, ranging from critical facility views to irrelevant content. To structure our analysis, we categorize images into five main facility types: \textit{Room Interior}, \textit{Indoor Facility}, \textit{Outdoor Facility}, \textit{Accommodation Exterior}, and \textit{Irrelevant}. To capture more specific functional contexts, images are additionally annotated with finer-grained sub-facility labels; the full taxonomy is provided in the Appendix~\ref{app:sub_categories}.

We define an image as \textit{informative} if it provides quantifiable visual cues along four fundamental dimensions: \textbf{Spatial Legibility (SL)}, \textbf{Activity Affordance (AA)}, \textbf{Contextual Openness (CO)}, and \textbf{Geometric Completeness (GC)}. Figure~\ref{fig:informativeness_axis} illustrates the characteristic visual patterns corresponding to each dimension. The applicability of these dimensions depends on the facility type, and Table~\ref{tab:info-factors} specifies which dimensions serve as valid evaluative criteria for each category.

Beyond these dimensions, \textit{Room Interior} images are additionally annotated with two semantic attributes—view type and room type—to capture preferences not fully represented by geometric or functional cues. Conversely, the \textit{Irrelevant} category contains images lacking decision-relevant visual evidence and is excluded from further evaluation.

\begin{table}[t]
\centering
\setlength{\tabcolsep}{4pt}
\begin{tabular}{lcccc}
\hline
\textbf{Facility Type} & \textbf{SL} & \textbf{AA} & \textbf{CO} & \textbf{GC} \\
\hline
Room Interior          & $\bullet$ & $\bullet$ &            &            \\
Indoor Facility        & $\bullet$ & $\bullet$ &            &            \\
Outdoor Facility       &           & $\bullet$ & $\bullet$  &            \\
Accommodation Exterior &           &           & $\bullet$  & $\bullet$  \\
\hline
\end{tabular}
\caption{
Facility types and applicable informativeness dimensions
(SL: Spatial Legibility; AA: Activity Affordance;
CO: Contextual Openness; GC: Geometric Completeness).
}
\label{tab:info-factors}
\end{table}

\subsection{Axis Definitions}

We define the four axes as quantifiable prediction targets to measure visual utility.

\paragraph{Spatial Legibility.} Defined as the count of distinct planar surfaces (floor, walls, ceiling), this metric serves as a proxy for \textbf{spatial comprehension}, distinguishing ambiguous close-ups from structural views that reveal room volume~\citep{oliva2001modeling}.

\paragraph{Activity Affordance.} We quantify \emph{meaningful components}—functional objects that explicitly afford guest activities (e.g., desks, seating, storage surfaces)—to capture the space’s \textbf{functional habitability} while filtering out purely decorative elements~\citep{Greene2016Function}.

\paragraph{Contextual Openness.} Measured as the ratio of non-facility elements (sky, nature, background structures), this metric assesses \textbf{contextual balance}, identifying overly occluded views or excessively distant shots that hinder environmental interpretation~\citep{CuestaValino2023Aesthetics}.

\paragraph{Geometric Completeness.} Approximating a building as a dominant cuboid, we assess the visibility of its three canonical faces—front, side, and roof—to evaluate \textbf{geometric integrity} and the perceptibility of its 3D form~\citep{SanchezTorres2024Visual}.

\medskip
For \textit{Room Interior} images, we supplement these structural axes with two semantic attributes—\textbf{View Type} and \textbf{Room Type}—which capture domain-specific preferences essential for booking decisions.

\section{Hospitality-VQA Dataset}
\label{sec:dataset}

\begin{figure}[t]
\centering
\fbox{%
\begin{minipage}{0.95\columnwidth}
\textbf{Hospitality-VQA Annotation Schema}

\vspace{4pt}
\textbf{(1) Hierarchical Labels} \\
\hspace*{1.2em}\texttt{main}: \textit{Primary facility category} \\
\hspace*{1.2em}\texttt{sub} \ : \textit{Fine-grained sub-facility}

\vspace{6pt}
\textbf{(2) Informativeness Axes} \\
\hspace*{1.2em}\texttt{SL}, \texttt{AA}, \texttt{CO}, \texttt{GC} \\
\hspace*{1.2em}\textit{*Mapped based on Table~\ref{tab:info-factors}}
\end{minipage}%
}
\caption{The formal annotation schema used in Hospitality-VQA. We record hierarchical facility labels and quantify visual utility across the four informativeness dimensions.}
\label{fig:annotation-schema}
\end{figure}

\begin{figure*}[t]
    \centering
    \includegraphics[width=\textwidth]{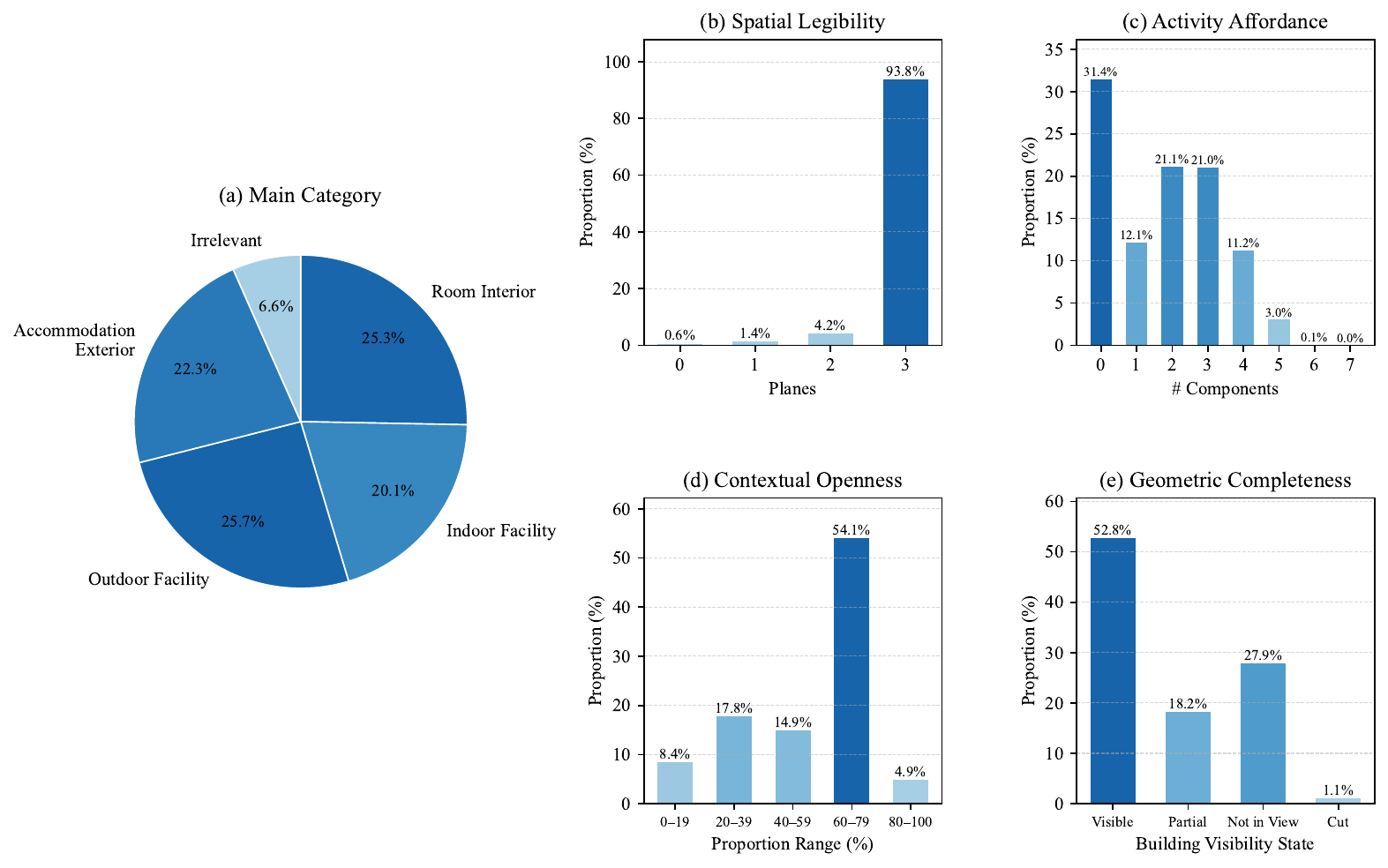}
    \caption{
    Dataset statistics of Hospitality-VQA. (a) Distribution of main facility categories. (b--e) Distributions of the four informativeness axes, reflecting characteristic properties of professionally curated hospitality listing images.
    }
    \label{fig:dataset-stats}
\end{figure*}

To translate the informativeness framework into a measurable benchmark, we introduce \textbf{Hospitality-VQA}. As existing VQA datasets lack the hospitality-specific imagery and annotations aligned with the four informativeness dimensions, they are ill-suited for evaluating decision-oriented visual reasoning. To address this gap, our dataset provides expert-annotated supervision explicitly tailored to these axes. The following subsections detail our pipeline for image collection, hierarchical annotation, and the derivation of instruction–answer pairs.

\subsection{Data Collection}

A total of 5,000 hospitality images were collected from \texttt{nol.yanolja.com} through random sampling of listing pages. Because the pool was not pre-filtered by facility type, categorization into facility types and relevance labels was performed during annotation (Section~\ref{sec:info}).

\subsection{Data Annotation}

Annotation was conducted by five annotators who were instructed in the Informativeness Framework defined in Section~\ref{sec:info}. Figure~\ref{fig:annotation-schema} summarizes the annotation schema used throughout dataset construction. A pilot round was conducted prior to the main annotation phase to calibrate labeling practices and align annotators' interpretations. All 5,000 images were then independently labeled by all annotators.

For quality control, we adopted a strict consensus protocol. Labels with high agreement (at least 4 out of 5 annotators)—covering 86.4\% of all annotations—were accepted as ground truth. Cases with lower agreement were flagged and resolved through consensus discussion, producing finalized facility-type and axis annotations for every image.

For VLM assessment, each label is converted into a concise instruction–answer pair using fixed templates specifically defined for each facility type and axis. These templates are designed to ensure consistent and scalable evaluation by mapping classification targets into a VQA format while controlling variation in question phrasing. Only the templates applicable to an image’s facility type are instantiated, and example templates are shown in Appendix~\ref{app:templates}.

\subsection{Dataset Analysis}
\label{sec:dataset-analysis}

Across the 5,000 collected images, a total of 19,729 QA pairs are generated by applying the fixed templates to the facility-type and axis-level annotations.
Figure~\ref{fig:dataset-stats} summarizes the overall distributions of facility categories and informativeness annotations.
As shown in Fig.~\ref{fig:dataset-stats}a, the main facility categories are relatively well balanced, each accounting for roughly a quarter of the dataset.

Figures~\ref{fig:dataset-stats}b--e report the distributions of the four informativeness dimensions.
Spatial Legibility (Fig.~\ref{fig:dataset-stats}b) and Activity Affordance (Fig.~\ref{fig:dataset-stats}c) are summarized as discrete counts reflecting visible planes and meaningful components, respectively, while Contextual Openness (Fig.~\ref{fig:dataset-stats}d) and Geometric Completeness (Fig.~\ref{fig:dataset-stats}e) are reported using predefined categorical bins.

Across these informativeness axes, we observe skewed distributions toward higher levels of visual informativeness. Such tendencies are characteristic of official listing imagery provided by hospitality platforms, which typically relies on professional photography to enhance spatial clarity, contextual visibility, and overall visual appeal for potential guests.
Unlike user-generated review photos, these images are intentionally composed to reveal room structure, spatial volume, and surrounding context.
As a result, the observed distributions reflect the visual evidence that users commonly encounter during actual booking decisions, supporting the ecological validity of our benchmark.
This structured coverage enables models to be evaluated not only on generic scene understanding, but also on the decision-relevant visual properties that matter in hospitality settings.

In Section~\ref{sec:experiments}, we use this dataset to benchmark several state-of-the-art VLMs and analyze their performance across facility types and informativeness axes.
\begin{table*}[t]
\centering
\setlength{\tabcolsep}{3.5pt} 
\renewcommand{\arraystretch}{1.08} 
\begin{tabular}{l|cc|cccc|cc}
\hline
\textbf{Model} &
\multicolumn{2}{c|}{\textbf{Facility}} &
\multicolumn{6}{c}{\textbf{Informativeness}} \\
\cline{2-9}
 & Main & Main\&Sub & SL & AA & CO & GC & Room & View \\
\hline
Gemini 2.5 Pro              & 90.66 & 75.00 & 11.51 &  9.43 & 46.81 &  7.35 & 80.65 & 50.00 \\
GPT-5                       & \underline{92.33} & 82.55 & 46.76 & 18.87 & 31.91 & 20.59 & \underline{83.87} & 64.10 \\
GPT-4o-mini                 & \underline{92.33} & \underline{84.91} & \textbf{97.12} & 38.21 & 56.03 &  8.82 & 70.97 & \textbf{79.49} \\
\hline
GLM-4.1V-9B-Thinking        & \textbf{93.66} & 79.25 & 89.21 & 35.85 & \underline{57.45} & 16.18 & 61.29 & 57.69 \\
LLaVA-NeXT-7B               & 73.33 & 53.77 & 94.24 &  8.02 & 19.86 &  5.88 & 25.81 & \textbf{79.49} \\
Gemma-3-12B                 & 92.00 & 82.08 & 86.33 & 15.09 & 55.32 & 22.06 & 72.19 & 43.59 \\
Qwen2.5-VL-3B               & 64.66 & 44.34 & 68.35 & 19.34 & 39.72 &  1.47 & 41.94 & 70.51 \\
Qwen2.5-VL-3B Finetuned     & 86.66 & 81.13 & \underline{94.96} & \underline{42.92} & \underline{57.45} & \underline{26.47} & 80.65 & \underline{76.92} \\
Qwen2.5-VL-7B               & 78.66 & 64.15 & 43.88 & 25.94 & 48.94 &  5.88 & 25.81 & 69.23 \\
\textbf{Qwen2.5-VL-7B Finetuned} & 92.00 & \textbf{85.37} & \textbf{97.12} & \textbf{44.34} & \textbf{67.37} & \textbf{32.35} & \textbf{87.10} & 74.36 \\
\hline
\end{tabular}
\caption{Comparison of VLM performance across facility types and informativeness categories.
Best in each column is in \textbf{bold} and second-best is \underline{underlined}.}
\label{tab:vlm_full}
\end{table*}

\section{Experiments}
\label{sec:experiments}

We evaluate a range of general-purpose Vision--Language Models (VLMs) on Hospitality-VQA to examine how well they capture the domain-specific informativeness axes introduced in Section~\ref{sec:info}.

\subsection{Experimental Setup}

\paragraph{Data split.}
Hospitality-VQA contains 5,000 labeled accommodation images.
We reserve 300 images for evaluation.
The remaining 4,700 images are used for training.
The evaluation split is sampled to preserve the overall distribution of facility types and informativeness factors, with class proportions matched within a 5\% margin relative to the full dataset.

\paragraph{Models.}
We evaluate eight vision--language models that span both commercial APIs and open-weight systems: GPT-5~\citep{openai2025gpt5}, GPT-4o-mini~\citep{hurst2024gpt4o}, Gemini 2.5 Pro~\citep{comanici2025gemini}, GLM-4.1V-9B-Thinking~\citep{hong2025glm}, Qwen2.5-VL-3B and Qwen2.5-VL-7B~\citep{bai2025qwen25vltechnicalreport}, LLaVA-NeXT-7B~\citep{li2024llavanext-strong}, and Gemma-3-12B~\citep{team2025gemma}. 
The proprietary models (GPT-5, GPT-4o-mini, Gemini 2.5 Pro, and GLM-4.1V-9B-Thinking) serve as strong general-purpose assistants that have been optimized for broad, web-scale multimodal use, whereas the open-weight models (Qwen2.5-VL-3B, Qwen2.5-VL-7B, LLaVA-NeXT-7B, and Gemma-3-12B) provide instruction-tuned checkpoints with varying capacities and training pipelines that are accessible for research and adaptation.
This combination allows us to examine how both deployment setting and model family affect performance on hospitality-oriented VQA.

Beyond zero-shot evaluation, we also derive task-adapted variants of Qwen2.5-VL-3B and Qwen2.5-VL-7B by applying LoRA fine-tuning~\citep{hu2022lora} on Hospitality-VQA.
In this configuration, the models are trained to predict the discrete axis labels in our framework from an image--prompt pair, aligning their outputs with our informativeness-oriented, classification-style supervision rather than generic captioning or open-ended generation.

\paragraph{Tasks and metrics.} To align with real-world hospitality applications (e.g., booking platforms) that require discrete, interpretable attributes rather than free-form text, we formulate all tasks as classification problems. We evaluate six core tasks—main facility type, main+sub facility type, visible plane count, meaningful component count, discretized scenery proportion, and building-face visibility—along with two auxiliary interior attributes: room and view type.

Models are prompted with a natural-language instruction template and must output a single categorical label. We report exact-match accuracy, reflecting the binary nature of practical decision-making; predictions that fail to map to a valid label are strictly penalized, mirroring real-world failure modes in attribute extraction systems. For API-based models, we use deterministic decoding (temperature = 0).

\subsection{Overall Results}
\label{sec:overall_results}

Table~\ref{tab:vlm_full} reports accuracy across facility recognition and all informativeness-related tasks. We summarize the results by (i) task difficulty across axes, (ii) model-family trends, and (iii) the effect of domain adaptation.

\subsubsection{Task Difficulty Across Axes}
\label{sec:task_difficulty_axes}

Table~\ref{tab:vlm_full} shows that \textit{main} facility classification transfers well across most evaluated VLMs, with several models exceeding 90\% accuracy. In contrast, \textit{main\&sub} recognition is consistently lower, indicating that fine-grained sub-category prediction is more demanding than coarse scene categorization under the same prompting and evaluation protocol.

Axis-level tasks exhibit a sharper drop in performance than facility recognition. Across models, Spatial Legibility (SL) is generally more stable than the other informativeness axes, whereas Activity Affordance (AA) and Geometric Completeness (GC) are notably weaker for many models. Contextual Openness (CO) falls between these extremes but still remains substantially below facility recognition performance, suggesting that decision-relevant attributes are not reliably recovered from generic multimodal capabilities alone.

Room and view attributes for \textit{Room Interior} show additional variability across models. While some models achieve strong accuracy on these auxiliary tasks, others lag despite high facility recognition, reinforcing that success on global categorization does not guarantee robust prediction of hospitality-relevant fine-grained attributes.

\subsubsection{Model Family Trends}
\label{sec:model_family_trends}

Model families show broadly similar behavior on coarse facility recognition but diverge more on axis-level prediction. Several proprietary models achieve high accuracy on \textit{main} facility classification, and some open-weight models also reach comparable levels, indicating that recognizing the overall facility category is not the primary bottleneck in this benchmark.

Differences become more pronounced for informativeness axes. For instance, GPT-4o-mini attains very high SL accuracy (97.12), yet AA and GC remain much lower (38.21 and 8.82). A similar pattern appears in multiple open-weight baselines (e.g., Qwen2.5-VL-7B: SL 43.88 vs.\ AA 25.94 and GC 5.88), where axis-level prediction does not track facility recognition. These results suggest that axis performance reflects additional reasoning requirements beyond generic scene labeling.

We avoid attributing these gaps to a single cause, as controlled ablations over training data, vision encoders, and instruction-tuning procedures are outside the scope of this work. Nonetheless, the consistent separation between facility recognition and axis-level performance across both proprietary and open-weight systems motivates explicit domain-grounded supervision for decision-oriented attributes.

\subsubsection{Effect of Domain Adaptation}
\label{sec:domain_adaptation}

Domain adaptation via LoRA fine-tuning~\citep{hu2022lora} consistently improves Qwen2.5-VL models across all evaluated tasks. Table~\ref{tab:adapt_gain} reports absolute gains (Finetuned$-$Base) computed from Table~\ref{tab:vlm_full}. Improvements are observed for both coarse facility recognition and fine-grained facility prediction, with particularly large gains on \textit{main\&sub} classification.

Gains are also evident on informativeness axes, which are challenging in the zero-shot setting. Notably, both model sizes improve on AA, CO, and GC, while the 7B model shows a pronounced increase on SL. Interior attributes benefit as well: room type accuracy increases substantially for both models, whereas view type shows smaller but consistent gains. Overall, these results indicate that axis-aligned supervision in Hospitality-VQA provides an effective signal for aligning VLM outputs with decision-oriented hospitality attributes under a strict label-matching evaluation.

\begin{table}[t]
\centering
\setlength{\tabcolsep}{5pt}
\renewcommand{\arraystretch}{1.05}
\begin{tabular}{lcc}
\hline
\textbf{Task} & \textbf{3B ($\Delta$ Acc)} & \textbf{7B ($\Delta$ Acc)} \\
\hline
Main      & +22.00 & +13.34 \\
Main\&Sub & +36.79 & +21.22 \\
SL        & +26.61 & +53.24 \\
AA        & +23.58 & +18.40 \\
CO        & +17.73 & +18.43 \\
GC        & +25.00 & +26.47 \\
Room      & +38.71 & +61.29 \\
View      & +6.41  & +5.13  \\
\hline
\end{tabular}
\caption{Absolute accuracy gains (\%) from domain adaptation via LoRA fine-tuning for Qwen2.5-VL models (Finetuned$-$Base).}
\label{tab:adapt_gain}
\end{table}

\section{Conclusion}

This work addressed the gap between general-purpose visual understanding and the kinds of fine-grained, decision-relevant reasoning required in the hospitality domain. 
While images play a central role in shaping guest expectations and booking decisions, existing multimodal systems lack the structured grounding necessary to interpret the spatial, functional, and view-related cues that matter in real domain use cases, just interpreting surface-level visual scenes. 

To bridge this gap, we introduced \emph{Hospitality Informativeness}, a domain-grounded framework that formalizes four fundamental visual axes—spatial legibility, activity affordance, contextual openness, and geometric completeness, whom are interpretable and measurable.
Building on this framework, we constructed \textbf{Hospitality-VQA}, a decision-centric VQA benchmark designed to elicit and evaluate the kinds of visual evidence that influence guest perception across diverse facility types.
E.g., whether models capture layout, functional components, scenery, and exterior visibility that matter for booking decisions.
Together, these contributions provide the first structured basis for measuring how well VLMs interpret hospitality imagery beyond generic scene recognition.

Our empirical study revealed that state-of-the-art general-purpose VLMs struggle with the fine-grained informativeness reasoning that the hospitality domain demands. 
However, we also showed that lightweight domain adaptation using our dataset leads to consistent and measurable improvements, highlighting both the challenge of the task and the value of the benchmark as a foundation for future model development.

\paragraph{Future Directions}
Looking ahead, Hospitality-VQA and the hospitality informativeness framework open several research directions, including domain-aware representation learning, prompt optimization, and test-time reasoning strategies. 
A particularly promising extension is modeling \emph{human-preferred accommodation attractiveness}, as user impressions are often shaped by images. 
This line of work carries clear practical value: \textbf{B2C} applications include displaying more appealing images to improve user experience and booking rates, while \textbf{B2B} applications involve curating and ranking property images based on user appeal. 
We hope our benchmark provides a foundation for future advances in hospitality-aware multimodal intelligence that benefits both users and service providers. 

\section*{Limitations}

This work has several limitations. 
First, Hospitality-VQA focuses on static images collected from a specific set of hotels and platforms, and the proposed informativeness axes represent a pragmatic but necessarily incomplete abstraction of real-world user information needs. 
In particular, while our framework emphasizes functional, spatial, and contextual visual cues, it does not explicitly capture aesthetic qualities such as visual style, ambiance, or emotional appeal, which can also influence user preferences in hospitality settings.

Second, our study does not model additional modalities or contextual factors commonly involved in accommodation decisions, such as textual reviews, pricing information, temporal media (e.g., videos), or personalized user preferences. 
As a result, the evaluation is limited to image-based visual reasoning under a controlled decision setting.

Third, all model evaluations are conducted under a single annotation protocol and question formulation. 
We do not assess the robustness of the reported results under alternative labeling schemes, prompt designs, or downstream task definitions.

Finally, although the dataset contains 5,000 annotated images in total, quantitative evaluation is performed on a held-out subset of 300 images. 
This relatively small evaluation set may limit statistical power and reduce sensitivity to rare or long-tail cases.

\section*{Acknowledgments}
The views and conclusions expressed in this paper are those of the authors and should not be interpreted as representing the official views, policies, or products of their affiliated organization.

\clearpage


\bibliography{custom}

@article{CuestaValino2023Aesthetics,
  title   = {The effects of the aesthetics and composition of hotels’ digital photo images on online booking decisions},
  author  = {Cuesta-Vali{\~n}o, Pedro and Kazakov, Sergey and Guti{\'e}rrez-Rodr{\'i}guez, Pablo and Lima Rua, Orlando},
  journal = {Humanities and Social Sciences Communications},
  volume  = {10},
  pages   = {59},
  year    = {2023},
  doi     = {10.1057/s41599-023-01529-w}
}

@article{SanchezTorres2024Visual,
  title   = {Visual photography’s influences on hotel selection: an analysis using e-booking as a comparative platform},
  author  = {S{\'a}nchez-Torres, Javier A. and Palacio-L{\'o}pez, Sandra-Milena and Hernandez-Fernandez, Yuri and Arroyo-Ca{\~n}ada, Francisco J. and Argila-Irurita, Ana},
  journal = {International Journal of Electronic Customer Relationship Management},
  volume  = {14},
  number  = {2},
  pages   = {128--142},
  year    = {2024},
  doi     = {10.1504/IJECRM.2023.10061227}
}

@article{comanici2025gemini,
  title={Gemini 2.5: Pushing the frontier with advanced reasoning, multimodality, long context, and next generation agentic capabilities},
  author={Comanici, Gheorghe and Bieber, Eric and Schaekermann, Mike and Pasupat, Ice and Sachdeva, Noveen and Dhillon, Inderjit and Blistein, Marcel and Ram, Ori and Zhang, Dan and Rosen, Evan and others},
  journal={arXiv preprint arXiv:2507.06261},
  year={2025}
}

@article{bai2025qwen25vltechnicalreport,
  title={{Qwen2.5-VL Technical Report}},
  author={Bai, Shuai and Chen, Keqin and Liu, Xuejing and Wang, Jialin and Ge, Wenbin and Song, Sibo and others},
  journal={arXiv preprint arXiv:2502.13923},
  year={2025}
}

@article{hurst2024gpt4o,
  title={{GPT-4o System Card}},
  author={Hurst, Aaron and others},
  journal={arXiv preprint arXiv:2410.21276},
  year={2024}
}

@inproceedings{trabelsi2025product,
  title={What Matters when Building Vision Language Models for Product Image Analysis?},
  author={Trabelsi, Ameni and Zontak, Maria and Qian, Yiming and Jackson, Brian and Khan, Suleiman and Batur, Umit},
  booktitle={Proceedings of the IEEE/CVF Winter Conference on Applications of Computer Vision Workshops (WACV Workshops)},
  year={2025}
}

@inproceedings{goyal2017making,
  title={Making the V in VQA Matter: Elevating the Role of Image Understanding in Visual Question Answering},
  author={Goyal, Yash and Khot, Tejas and Summers-Stay, Douglas and Batra, Dhruv and Parikh, Devi},
  booktitle={Proceedings of the IEEE Conference on Computer Vision and Pattern Recognition (CVPR)},
  year={2017}
}

@inproceedings{hudson2019gqa,
  title={GQA: A new dataset for real-world visual reasoning and compositional question answering},
  author={Hudson, Drew A and Manning, Christopher D},
  booktitle={Proceedings of the IEEE Conference on Computer Vision and Pattern Recognition (CVPR)},
  year={2019}
}

@inproceedings{das2018embodied,
  title={Embodied Question Answering},
  author={Das, Abhishek and Datta, Samyak and Gkioxari, Georgia and Lee, Stefan and Parikh, Devi and Batra, Dhruv},
  booktitle={Proceedings of the IEEE Conference on Computer Vision and Pattern Recognition (CVPR)},
  year={2018}
}

@inproceedings{gurari2018vizwiz,
  title={VizWiz: Nearly Real-Time Answers to Visual Questions},
  author={Gurari, Danna and Li, Quchen and Stangl, Anthony J and Guo, Yongsen and Lin, Chuan-He and Grauman, Kristen and Luo, Jiebo and Bigham, Jeffrey P},
  booktitle={Proceedings of the IEEE Conference on Computer Vision and Pattern Recognition (CVPR)},
  year={2018}
}

@article{wakefield1996servicescape,
    author = {Wakefield, Kirk L. and Blodgett, Jeffrey G.},
    title = {The effect of the servicescape on customers’ behavioral intentions in leisure service settings},
    journal = {Journal of Services Marketing},
    volume = {10},
    number = {6},
    pages = {45-61},
    year = {1996},
    month = {12},
    issn = {0887-6045},
    doi = {10.1108/08876049610148594},
    url = {https://doi-org-ssl.access.yonsei.ac.kr/10.1108/08876049610148594},
    eprint = {https://www-emerald-com-ssl.access.yonsei.ac.kr/jsm/article-pdf/10/6/45/1710229/08876049610148594.pdf},
}

@article{Greene2016Function,
  title   = {Visual Scenes Are Categorized by Function},
  author  = {Greene, Michelle R. and Baldassano, Christopher and Esteva, Andre and Beck, Diane M. and Fei-Fei, Li},
  journal = {Journal of Experimental Psychology: General},
  volume  = {145},
  number  = {1},
  pages   = {82--94},
  year    = {2016},
  doi     = {10.1037/xge0000129}
}

@article{zhang2022makes,
  title={What makes a good image? Airbnb demand analytics leveraging interpretable image features},
  author={Zhang, Shunyuan and Lee, Dokyun and Singh, Param Vir and Srinivasan, Kannan},
  journal={Management Science},
  volume={68},
  number={8},
  pages={5644--5666},
  year={2022},
  publisher={INFORMS}
}

@article{li2013determinants,
  title={Determinants of customer satisfaction in the hotel industry: An application of online review analysis},
  author={Li, Huiying and Ye, Qiang and Law, Rob},
  journal={Asia Pacific journal of tourism research},
  volume={18},
  number={7},
  pages={784--802},
  year={2013},
  publisher={Taylor \& Francis}
}

@article{xiang2015can,
  title={What can big data and text analytics tell us about hotel guest experience and satisfaction?},
  author={Xiang, Zheng and Schwartz, Zvi and Gerdes Jr, John H and Uysal, Muzaffer},
  journal={International journal of hospitality management},
  volume={44},
  pages={120--130},
  year={2015},
  publisher={Elsevier}
}

@article{he2023image,
  title={Image features and demand in the sharing economy: A study of Airbnb},
  author={He, Jiaxiu and Li, Bingqing and Wang, Xin Shane},
  journal={International Journal of Research in Marketing},
  volume={40},
  number={4},
  pages={760--780},
  year={2023},
  publisher={Elsevier}
}

@article{ren2021large,
  title={Large-scale comparative analyses of hotel photo content posted by managers and customers to review platforms based on deep learning: implications for hospitality marketers},
  author={Ren, Meng and Vu, Huy Quan and Li, Gang and Law, Rob},
  journal={Journal of Hospitality Marketing \& Management},
  volume={30},
  number={1},
  pages={96--119},
  year={2021},
  publisher={Taylor \& Francis}
}

@article{guidotti2025discovering,
  title={Discovering sentiment insights: streamlining tourism review analysis with Large Language Models},
  author={Guidotti, Dario and Pandolfo, Laura and Pulina, Luca},
  journal={Information Technology \& Tourism},
  volume={27},
  number={1},
  pages={227--261},
  year={2025},
  publisher={Springer}
}

@article{hong2025glm,
  title={GLM-4.1 V-Thinking: Towards Versatile Multimodal Reasoning with Scalable Reinforcement Learning},
  author={Hong, Wenyi and Yu, Wenmeng and Gu, Xiaotao and Wang, Guo and Gan, Guobing and Tang, Haomiao and Cheng, Jiale and Qi, Ji and Ji, Junhui and Pan, Lihang and others},
  journal={arXiv preprint arXiv:2507.01006},
  year={2025}
}

@article{team2025gemma,
  title={Gemma 3 technical report},
  author={Team, Gemma and Kamath, Aishwarya and Ferret, Johan and Pathak, Shreya and Vieillard, Nino and Merhej, Ramona and Perrin, Sarah and Matejovicova, Tatiana and Ram{\'e}, Alexandre and Rivi{\`e}re, Morgane and others},
  journal={arXiv preprint arXiv:2503.19786},
  year={2025}
}

@article{hu2022lora,
  title={Lora: Low-rank adaptation of large language models.},
  author={Hu, Edward J and Shen, Yelong and Wallis, Phillip and Allen-Zhu, Zeyuan and Li, Yuanzhi and Wang, Shean and Wang, Lu and Chen, Weizhu and others},
  journal={ICLR},
  volume={1},
  number={2},
  pages={3},
  year={2022}
}

@misc{li2024llavanext-strong,
    title={LLaVA-NeXT: Stronger LLMs Supercharge Multimodal Capabilities in the Wild},
    url={https://llava-vl.github.io/blog/2024-05-10-llava-next-stronger-llms/},
    author={Li, Bo and Zhang, Kaichen and Zhang, Hao and Guo, Dong and Zhang, Renrui and Li, Feng and Zhang, Yuanhan and Liu, Ziwei and Li, Chunyuan},
    month={May},
    year={2024}
}

@misc{openai2025gpt5,
    title={GPT-5 System Card},
    url={https://cdn.openai.com/gpt-5-system-card.pdf},
    author={{OpenAI}},
    month={August},
    year={2025}
}

@article{oliva2001modeling,
  title={Modeling the shape of the scene: A holistic representation of the spatial envelope},
  author={Oliva, Aude and Torralba, Antonio},
  journal={International journal of computer vision},
  volume={42},
  number={3},
  pages={145--175},
  year={2001},
  publisher={Springer}
}

@article{wei2022chain,
  title={Chain-of-thought prompting elicits reasoning in large language models},
  author={Wei, Jason and Wang, Xuezhi and Schuurmans, Dale and Bosma, Maarten and Xia, Fei and Chi, Ed and Le, Quoc V and Zhou, Denny and others},
  journal={Advances in neural information processing systems},
  volume={35},
  pages={24824--24837},
  year={2022}
}

@article{tu2024towards,
  title={Towards generalist biomedical AI},
  author={Tu, Tao and Azizi, Shekoofeh and Driess, Danny and Schaekermann, Mike and Amin, Mohamed and Chang, Pi-Chuan and Carroll, Andrew and Lau, Charles and Tanno, Ryutaro and Ktena, Ira and others},
  journal={Nejm Ai},
  volume={1},
  number={3},
  pages={AIoa2300138},
  year={2024},
  publisher={Massachusetts Medical Society}
}

@article{askari2025hotelmatch,
  title={HotelMatch-LLM: Joint Multi-Task Training of Small and Large Language Models for Efficient Multimodal Hotel Retrieval},
  author={Askari, Arian and Stergiadis, Emmanouil and Gusev, Ilya and Beladev, Moran},
  journal={arXiv preprint arXiv:2506.07296},
  year={2025}
}

\clearpage
\appendix

\section{Detailed Facility Taxonomy}
\label{app:sub_categories}

To support consistent annotation and evaluation, we define a hierarchical facility taxonomy with clear operational criteria.
Images are first assigned to one of five main facility types based on accessibility and visible structural boundaries:
\textit{Room Interior}, \textit{Indoor Facility}, \textit{Outdoor Facility}, \textit{Accommodation Exterior}, and \textit{Irrelevant}.

\textbf{Room Interior} is restricted to private guest spaces.
In cases of spatial overlap (e.g., studio-type rooms), a fixed priority order is applied
(\textit{Bedroom} $>$ \textit{Kitchen} $>$ \textit{Bathroom} $>$ \textit{Living room})
to ensure unique assignment.
\textbf{Indoor} and \textbf{Outdoor Facilities} are distinguished by whether the space is fully enclosed, with outdoor facilities required to be the primary visual focus rather than part of a general landscape.
\textbf{Accommodation Exterior} is assigned only when the building itself constitutes the main subject with identifiable accommodation features.
Images lacking discernible hospitality context are grouped into the \textbf{Irrelevant} category, which functions as a noise class.

For finer-grained functional analysis, each main category (except \textit{Accommodation Exterior} and \textit{Irrelevant}) is further annotated with sub-facility labels, summarized in Table~\ref{tab:full-taxonomy}.
This granularity enables evaluation of whether models can recognize specific functional contexts relevant to hospitality decision-making.

\begin{table}[htbp]
\centering
\setlength{\tabcolsep}{4pt}
\renewcommand{\arraystretch}{1.1}
\begin{tabular}{@{}p{0.32\columnwidth}p{0.64\columnwidth}@{}}
\hline
\textbf{Main Category} & \textbf{Sub-category} \\
\hline
Room Interior &
Bedroom; Kitchen; Bathroom; Living room \\
\hline
Indoor Facility &
Guest lounge; Reception desk; Hallway; Restaurant \& Cafe; Indoor pool; Indoor parking lot; Other amenities \\
\hline
Outdoor Facility &
Outdoor pool \& Spa; Outdoor lounge \& BBQ area; Sports \& Recreation facility; Outdoor parking lot; Camping area \\
\hline
Accommodation Exterior & --- \\
\hline
Irrelevant & --- \\
\hline
\end{tabular}
\caption{Full taxonomy of hospitality facility classification and sub-category labels.}
\label{tab:full-taxonomy}
\end{table}

\section{Instruction-Answer Construction Template}
\label{app:templates}

This appendix provides details on how the expert-verified labels are mapped into instruction--answer pairs using our fixed templates.
As described in Section~\ref{sec:dataset-analysis}, these templates are designed to ensure consistency across the dataset by formatting classification targets into a standardized VQA format.

To maintain evaluation rigor, each template consists of a task-specific prompt and a constrained answer format. 
Figure~\ref{fig:template_general} illustrates the general structure of these templates. 
Representative examples for facility-type classification and informativeness axis evaluation are presented in Figures~\ref{fig:template_main} and~\ref{fig:template_gc}, respectively.

\begin{figure}[t]
\centering
\fbox{
\begin{minipage}{0.95\columnwidth}
\textbf{General Instruction--Answer Template Format}\\[4pt]

\textbf{Task} \\
Target classification or assessment objective.\\[3pt]

\textbf{Prompt} \\
Natural language instruction defining task semantics and decision rules.\\[3pt]

\textbf{Answer Format} \\
Strictly constrained output schema (e.g., class ID or fixed key--value pairs).\\[3pt]

\textbf{Answer} \\
Example output following the specified format.
\end{minipage}}
\caption{General structure of instruction--answer templates shared across all evaluation tasks.}
\label{fig:template_general}
\end{figure}

\begin{figure*}[b]
\centering
\fbox{
\parbox{0.95\linewidth}{
\textbf{Task: Facility Type (Main)}\\
\textbf{Prompt:} Your task is to classify given image.

Definitions and specific instructions for each category are as follows:

1.

Private accommodation room interior space for guest sleeping/living functions. Includes bedrooms, bathrooms, living rooms, kitchens, and photos taken from inside rooms. Shared areas or facilities do not belong to this category.

2.

Shared "indoor" facilities within accommodation (e.g. customer lounges, reception desks, corridors, restaurants/cafes, indoor pools, indoor parking, other amenities (gyms, indoor golf, saunas, convenience stores, seminar rooms etc.))

3.

Specific "outdoor" facilities that falls into following cases: outdoor pools/spas, outdoor lounges/garden/terrace/BBQ areas, outdoor sports/recreation facilities, outdoor parking, outdoor camping areas. Must be clearly identifiable as one of these facility types and be the image's primary focus, not part of general accommodation or landscape views. Exclude: overall building/accommodation views even if outdoor facilities are visible, pure nature shots without specific facilities.

4.

Image showing accommodation building exterior AS THE MAIN SUBJECT. Buidling must occupy significant portion of image with clear structural elements (walls, windows, roof) and typical accommodation features (guestroom windows, balconies, nearby amenities) to be identifiable as accommodation.
Only for photos that do NOT fall into categories 1, 2, 3, or 5 AND where the building itself is the primary focus, not background.
Excluded: no visible building, appears to be non-residential buildings due to lack of accommodation features, main accommodation building unclear among multiple scattered buildings, accommodation too small/distant to recognize (e.g. tiny in a wide drone shot).

5.

Image lacking any clues identifying them as prior 4 categories of accommodation. Includes pure nature shots, pet/person-focused shots without spatial clues, notice/text-based images (e.g., posters, receipts), and building exteriors not meeting criteria of 4. For close-up images, prefer classifying to 1-4 over 5 when possible, if there are any clues suggesting accommodation context, even if subtle (e.g., cushion seems to be on bed → 1, food seems to be in restaurant → 2).

ANSWER FORMAT

Output a single number: <1-5>

Do not include any explanation, spaces, or other characters.\\
\textbf{Answer:} 1
}}
\caption{Example of constructed instruction–answer template of main facility.}
\label{fig:template_main}
\end{figure*}

\begin{figure*}[!htbp]
\centering
\fbox{
\parbox{0.95\linewidth}{
\textbf{Task: Geometric Completeness}\\
\textbf{Prompt:} You are an expert image analyst specializing in architectural assessments.

Your task is to analyze the visible faces of the single most plausible and visually prominent lodging building in an image.

For the selected building, output a visibility status code (1–4) for each of these three faces:

- '1' = Front Facade

- '2' = Side Wall

- '3' = Roof

Apply the rules in this order for each face:

1. Assign 1 if the face is absent or not visible at all.

2. Assign 2 if a clear, identifiable portion of the face is cut off by the image's edges.

3. Assign 3 if the face is visible but significantly blocked by an external object.

4. Assign 4 if the face is clearly visible and unobstructed (roof only if distinct and unambiguous).

ANSWER FORMAT

Output exactly in this format, with no spaces or extra text:

'1': <1-4>, '2': <1-4>, '3': <1-4>

\textbf{Answer:} '1': 3, '2': 3, '3': 4
}}
\caption{Example of constructed instruction–answer template of geometric completeness.}
\label{fig:template_gc}
\end{figure*}



\clearpage

\section{Additional Experimental Details}
\label{app:experiments_details}

We provide implementation details to support the reproducibility of the results in Section~\ref{sec:experiments}. All fine-tuning experiments on open-weight models were conducted on a single NVIDIA RTX~4090 GPU using the \texttt{unsloth} framework.

\subsection{Training Setup}
Models were fine-tuned for two epochs using supervised learning on image--instruction pairs. We used the AdamW optimizer with a learning rate of $2\times10^{-5}$, 5\% linear warmup, and cosine decay. The effective batch size was 16 (batch size 2 per device with gradient accumulation of 8). Training was performed in bfloat16 precision with a maximum context length of 8{,}192 tokens. Weight decay and gradient clipping were not applied.

\subsection{LoRA Configuration}
We adopted Low-Rank Adaptation (LoRA)~\citep{hu2022lora} with rank $r=16$ and scaling factor $\alpha=32$. Adapters were inserted into the vision encoder and language decoder, covering the attention projections and MLP layers. LoRA dropout was set to 0, and all other model parameters were frozen.

\begin{table*}[ht!]
\centering
\setlength{\tabcolsep}{3.8pt}
\renewcommand{\arraystretch}{1.06}
\begin{tabular}{l|cc|cccc|cc}
\hline
\textbf{Model} &
\multicolumn{2}{c|}{\textbf{Facility}} &
\multicolumn{6}{c}{\textbf{Informativeness}} \\
\cline{2-9}
 & Main & Main\&Sub & SL & AA & CO & GC & Room & View \\
\hline
Qwen2.5-VL-3B                    & 64.66 & 44.34 & 68.35 & 19.34 & 39.72 &  1.47 & 41.94 & 70.51 \\
Qwen2.5-VL-3B FT (w/o CoT)       & 86.66 & 81.13 & 94.96 & 42.92 & 57.45 & 26.47 & 80.65 & \textbf{76.92} \\
Qwen2.5-VL-3B FT (w/ CoT)        & 85.66 & 73.58 & 93.53 & 34.91 & 63.38 & 27.94 & 64.52 & 70.51 \\
\hline
Qwen2.5-VL-7B                    & 78.66 & 64.15 & 43.88 & 25.94 & 48.94 &  5.88 & 25.81 & 69.23 \\
Qwen2.5-VL-7B FT (w/o CoT)       & \textbf{92.00} & \textbf{85.37} & \textbf{97.12} & \textbf{44.34} & \textbf{67.37} & \textbf{32.35} & \textbf{87.10} & 74.36 \\
Qwen2.5-VL-7B FT (w/ CoT)        & 91.33 & 83.02 & 94.24 & 42.45 & 59.57 & 26.47 & 83.87 & \textbf{76.92} \\
\hline
\end{tabular}
\caption{Comparison of VLM performance with and without CoT. Best in each column is highlighted in \textbf{bold}.}
\label{tab:vlm_cot}
\end{table*}

\section{CoT vs. No CoT}

We explicitly investigated the impact of incorporating Chain-of-Thought (CoT)~\citep{wei2022chain} reasoning during the supervised fine-tuning process. Table~\ref{tab:vlm_cot} presents a performance comparison between the base models, models fine-tuned with CoT supervision, and models fine-tuned with direct answers (w/o CoT).

In the 3B setting, CoT supervision provides small gains on a few attributes (e.g., \textit{Scenery} and \textit{Building Faces}), but these improvements are neither consistent across tasks nor robust across model scales. Overall, direct-answer supervision without CoT yields more reliable performance for our classification-oriented evaluation.


\end{document}